\documentclass{Interspeech2024}
\usepackage{graphicx}
\usepackage{verbatim}

\usepackage{arydshln}

\interspeechcameraready

\title{Non-Linear Inference Time Intervention: Improving LLM Truthfulness}

\name[affiliation=*{1,3}]{Jakub}{Hoscilowicz}
\name[affiliation=*{1}]{Adam}{Wiacek}
\name[affiliation={1,2}]{Jan}{Chojnacki}
\name[affiliation={1}]{Adam}{Cieslak}
\name[affiliation={1}]{Leszek}{Michon}
\name[]{Vitalii}{Urbanevych} 
\name[affiliation={3}]{Artur}{Janicki}

\address{
  $^1$Samsung R\&D Institute Poland, Warsaw, Poland\\
  $^2$University of Warsaw, Poland\\
  $^3$Warsaw University of Technology, Poland
}
\email{\{j.hoscilowic, a.wiacek2, a.cieslak, l.michon\}@samsung.com,\\ jr.chojnacki@uw.edu.pl, artur.janicki@pw.edu.pl\thanks{$^*$The first two authors made equal contribution.}}

\begin{document}

\maketitle
\
\begin{abstract}
In this work, we explore LLM's internal representation space to identify attention heads that contain the most truthful and accurate information. We further developed the Inference Time Intervention (ITI) framework, which lets bias LLM without the need for fine-tuning. The improvement manifests in introducing a non-linear multi-token probing and multi-token intervention: Non-Linear ITI (NL-ITI), which significantly enhances performance on evaluation benchmarks. NL-ITI is tested on diverse multiple-choice datasets, including TruthfulQA, on which we report over \SI{16}{\percent} relative MC1 (accuracy of model pointing to the correct answer) improvement with respect to the baseline ITI results. Moreover, we achieved a 10\% relative improvement over the recently released Truth Forest (TrFf) method that also focused on ITI improvement. 
\end{abstract}


\section{Introduction}
Large Language Models (LLMs) are a major achievement in the domain of artificial intelligence, particularly within natural language processing (NLP). Their capabilities span a wide array of applications, from generating human-like texts to understanding and processing complex language structures. However, the probabilistic nature of these models often gives rise to certain challenges, including the phenomena of hallucinations~\cite{manakul2023selfcheckgpt, li2023halueval} and the generation of toxic content~\cite{shaikh2023second}. LLM models trained on extensive datasets inadvertently absorb and repeat cultural, gender-based, racial, or ideological biases in their training dataset~\cite{taubenfeld2024systematic, yeh2023evaluating}.
These issues underscore the motivation behind the development of robust benchmarks and methodologies aimed at evaluating and enhancing the safety, fairness, and accuracy of LLM outputs. A comprehensive strategy is necessary that includes diversifying training datasets, developing algorithms to detect and neutralize bias, and implementing robust testing protocols for biased outputs. Recent developments suggest how to assess and mitigate the model's bias~\cite{zou2023representation, li2023inferencetime, hartvigsen2022toxigen, parrish-etal-2022-bbq, lin2022truthfulqa}.

Our work investigates the internal representation space of LLMs to identify and utilize the most informative attention heads for specific tasks. During inference, the activations of such heads are modified, thus refining LLM-generated content. Our primary contribution is a notable enhancement of the Inference Time Intervention (ITI) method~\cite{li2023inferencetime}, leading to higher performance on LLM benchmarks and better generalization capability. The improvements manifest in two distinct \\\\\\aspects: firstly, the introduction of non-linearity to the probing model, which facilitates a more effective identification of attention heads collecting the type of desired knowledge (e.g., truthfulness). Secondly, the employment of an expanded token context during interventions enables a more refined construction of the intervention vector, thereby directing attention heads more effectively toward truthfulness. This enhanced construction of the intervention vector is attributed to the observation that truthful knowledge is not solely concentrated in the vector corresponding to the final token, but is distributed across a broader context. We discuss how our framework can be used to bias LLM toward any abstract concept (truthfulness, correctness, toxicity-prevention). We present our advancements and their contribution to developing safer, more accurate, and ethically responsible LLM systems, demonstrating the potential of our approach for future AI applications.

\section{Related Work}
 Efforts to mitigate LLM biases have led to the development of diverse strategies, of which one of the most impactful is the Reinforcement Learning from Human Feedback (RLHF)~\cite{ouyang2022training}. It aligns models with human feedback, reducing bias by adjusting model behaviors based on human preferences. Such an approach, while effective, demands significant human labor~\cite{lee2023rlaif}, highlighting the need for novel, automated bias mitigation methods.

In contrast, approaches like ITI~\cite{li2023inferencetime}
act more directly by modifying the model's internal representations. It was noticed that the LLMs sometimes `know' they produce false statements~\cite{kadavath2022language}. This motivated the ITI authors to bias the model towards more truthful behavior.  This involves a two-step process, where attention heads are first evaluated for their accuracy, and then Mass Mean Shift vectors are applied to a subset of top-performing heads during inference. These vectors, calculated as the mean difference between activations for true and false answers, are pre-computed and attached to the model, enabling a precision improvement with minimal computational overhead. 

Recently, another work focusing on analyzing and improving the probing procedure of the internal representation space has been reported. Authors of~\cite{chen2024truth} introduced the Truth Forest (TrFr) employing multi-dimensional orthogonal probes. However still, they focused on optimizing the original ITI framework.


The increasing demand for LLMs in a practical text generation underscores the importance of model fairness and truthfulness. Benchmarks such as TruthfulQA~\cite{lin2022truthfulqa} have been introduced to evaluate model truthfulness across various domains, including health, law, finance, and politics. These benchmarks challenge models with questions designed to elicit imitative falsehoods, thereby testing the model's ability to maintain truthfulness across topics. Similarly, datasets like BBQ~\cite{parrish-etal-2022-bbq} and ToxiGen~\cite{hartvigsen2022toxigen} assess LLM fairness and ability to handle nuanced manifestations of hate speech.

Recent work has sparked an interest in LLM personality categorization and psychometrics. Contributions in this area include adjusting LLM personality traits~\cite{mao2023editing}, using the Myers-Briggs Type Indicator (MBTI) for evaluation~\cite{pan2023llms}, and simulating diverse personalities in models~\cite{huang2023revisiting}.
Our work builds upon these foundational efforts, aiming to contribute to the development of more ethical and unbiased LLM applications.

\section{Method}
As reported in~\cite{li2023inferencetime}, LLMs seem to preserve an internal representation of abstract concepts such as truth and honesty, even though they do not generate factual responses. Simple prompt stimulation may not be enough to access the full potential stored in the internal representations.
In~\cite{li2023inferencetime}, one uses labeled data (i.e. question-answer Q+A pairs from TruthfulQA split) to train a linear probing model, which identifies attention heads storing the truthful representations. For each such head, the truthful direction is calculated. During the inference, attention head activations are shifted in the truthful direction.

This method unfolds in two phases. Initially, a linear probing model is trained on representations returned by the attention heads for a given probing trainset. The assumption is that the higher the accuracy of the probing model, the higher the amount of desired knowledge (e.g., truthful). Mathematically, the probing operation $p_\theta$ may be described as:
\begin{equation}
\label{eq:linear_probing}
    p_\theta\left( x_l^h\right)=\textrm{sigmoid}\left(\left\langle \theta, x_l^h \right\rangle\right),
\end{equation}
where $\theta$ is a set of trainable parameters, and $x^h_l$ is an activation of token $x$ at head $h$ and layer $l$. In the process of probe training, $N$ question-answers pairs are concatenated and the activations that correspond to the last token $\left\{ (x^h_l, y)_i\right\}^N_{i=1}$ are collected (where $y$ is a binary label that indicates true or false answer). 

Note that the probes need to be trained on a labeled dataset reflecting the concept (here, we used TruthfulQA~\cite{lin2022truthfulqa}). For each concept (e.g., truthfulness, toxicity-prevention, personality adjustment), a different biasing dataset and probe training are necessary. 

Following \cite{li2023inferencetime}, the intervention is then given by:
\begin{equation}
\label{eq:ITI}
    x_{l+1} = x_l +\sum^H_{h=1}Q^h_l\left(\textrm{Att}^h_l\left( P^h_l x_l\right) + \alpha\sigma^h_l\theta^h_l\right),
\end{equation} 
where $x_l\in \mathbb{R}^{DH}$ is the $l$-th token high-dimensional embedding, $P^h_l\in\mathbb{R}^{D\times DH}$ is a mapping operator from token embedding to $D$-dimensional attention head space, and $\textrm{Att}^h_l$ operator connects information from other tokens and gives us activations seen in Equation \ref{eq:linear_probing}: $x^h_l = \textrm{Att}^h_l \left(P^h_lx_l \right)$. Finally, the last term in Equation \ref{eq:ITI} is the intervention term. It can be understood as follows: when calculating the next-token-prediction $x_{l+1}$ the residual stream (previous $x_l$ token and the weighted sum of activations) is modified by adding the biasing direction $\theta^h_l$ multiplied by its standard deviation (with respect to all of the Q+A pairs) and the \textit{intervention strength} $\alpha$.

To calculate the biasing direction for each head and layer, one takes an average of activations of the last token among the Q+A dataset:
\begin{equation}
\label{eq:biasing_direction}
    \theta^h_l = \frac{1}{N}\sum^N_{i=1}\left(x^h_l\right)_i, \quad h \in \textrm{Top Heads}.
\end{equation}
The biasing directions are appended only for a number of attention heads with the highest probing accuracy. Their number is controlled by a second hyperparameter $K$.

\section{Proposed improvement}
\label{section:Proposed improvement}
ITI~\cite{li2023inferencetime} uses attention head probing based on a logistic regression probing model, as shown in Equation~\ref{eq:linear_probing}. We think that it does not optimally capture the complexity of the concept representation in the activation space. 

Therefore, we propose improved probing and suggest using non-linear MultiLayer Perceptron (MLP) as the probe, changing Equation \ref{eq:linear_probing} to:
\begin{equation}
    p_\theta\left(x^h_l\right) = \textrm{MLP}\left(\left\langle \theta, x_l^h \right\rangle\right).
\end{equation}
 Improving probes' accuracy leads to a more appropriate choice of the top heads. The top heads are then used in the ITI procedure described in Equation~\ref{eq:ITI}. Non-linear probing has generally higher information capacity and is able to capture more of the inherent linguistic information in the representation~\cite{pimentel2020informationtheoretic,white2021nonlinear, hościłowicz2023use}.

Moreover, instead of using only the last token for probing training, we focus on the average optimal number of last tokens. Increasing the information capacity of the probing model can be naturally followed by providing more information encoded in multiple tokens provided to the MLP.

This modification can be mathematically expressed as a change in which one collects the training dataset:
\begin{equation}
\label{eq:multi-token probing}
    \left\{\left( x^h_l,y\right)_i\right\}^N_{i=1} \xrightarrow{\textrm{multi-token}}
    \left\{\left( \left\langle x^h_l \right \rangle_\tau,y\right)_i\right\}^N_{i=1},
\end{equation}
where instead of taking just the last token, on which the activation $x^h_l$ are calculated, we take an average of $\tau$ last tokens. This optimum is found experimentally and described in detail in Section~\ref{section:Evaluation results}. 

Similarly, extending the number of tokens used in intervention is also relevant. This is our second improvement to the framework.

In particular, during the inference, at each attention head, we add biasing directions corresponding to the mean of the last $\rho$ tokens in a Q+A pair. To calculate the biasing direction vector we average over all Q+A pairs as in Equation~\ref{eq:biasing_direction}:
\begin{equation}
\label{eq:multi-token biasing_direction}
    \theta^h_l = \frac{1}{N}\sum^N_{i=1}\left(\left\langle x^h_l \right\rangle_\rho\right)_i, \quad h \in \textrm{Top Heads}.
\end{equation}
As before, the number of tokens $\rho$, over which the activations are averaged, needs to be found empirically.

From now on, we will call this framework \textit{Non-Linear-Inference Time Intervention} (NL-ITI). 

\section{Experiments}
\label{section:Evaluation results}
 We verify how token addition and probing non-linearity affect evaluation metrics. We discuss the generalization capabilities of NL-ITI and show the performance improvement on diverse reasoning tests.

\begin{figure*}[ht!]
  \centering
  \includegraphics[width=0.75\textwidth]{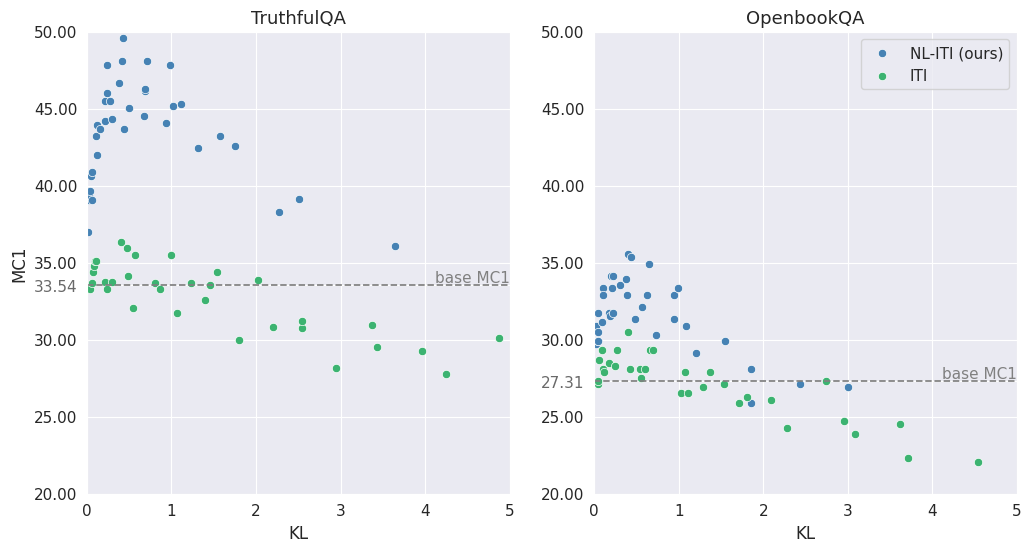}
  \caption{How  MC1 correlates with KL divergence. The results were collected for ITI and NL-ITI using different hyperparameter sets ($\alpha$, heads to intervene) for TruthfulQA and OpenBookQA datasets. On each benchmark, baseline LLaMA-2-7B performance is shown.}
 \label{fig:MC1(KL)}
\end{figure*}
\subsection{Evaluation}
Throughout this work, all experiments were performed on LLaMA-2-7B~\cite{touvron2023llama} model. This open-source LLM has been used extensively in previous work on AI safety. Particularly, it let us to directly compare NL-ITI to related methods~\cite{li2023inferencetime, chen2024truth}.

Figure \ref{fig:MC1(KL)} shows how ITI and NL-ITI compare on two benchmarks. Each point corresponds to a different hyperparameter set, reflected in different Kullback-Leibler divergence (KL) calculated with respect to OpenWebText~\cite{Gokaslan2019OpenWeb} distribution. At each test-point, NL-ITI outperforms ITI. NL-ITI creates a peak in $\textrm{KL}\sim 0.5$, the correlation is also non-linear around this region.

In~\cite{li2023inferencetime}, the authors report MC1 and MC2 scores introduced in TruthfulQA~\cite{lin2022truthfulqa} benchmark and generative evaluation methods.
MC1 and MC2 can be understood as the accuracy at which the model predicts the correct answer for a given question, if the model output was to be restricted only to generate one of the (correct or otherwise) answers specified in the dataset. MC1 is applicable for only single-correct answer datasets, while MC2 is well-defined also for multiple-correct datasets. For the implementation of MC1 and MC2 scores we reference \cite{lin2022truthfulqa} code and our GitHub repository\footnote{https://github.com/Samsung/NL-ITI}.
 We proceed with MC1 and MC2 score evaluation, as they are more reliable and replicable independently of the Judge LLM Models. MC-scores do not depend on the underlying judging model reasoning capabilities, since labeled data is used in the accuracy calculation. Moreover, this allows us for more direct comparisons with other approaches~\cite{li2023inferencetime, chen2024truth}, as the evaluations with closed-source GPT-4 may depend on the OpenAI software updates.

Similarly to~\cite{li2023inferencetime}, Cross Entropy (CE) and KL divergence are calculated to see how much the intervention result diverges from the original LLaMA-2-7B~\cite{touvron2023llama} token distribution. For both metrics, lower values correspond to less change in the model's behavior. Very large CE and KL values suggest intervention procedure changed the output token distribution in a major way. One could imagine that such a change could negatively affect LLM's language comprehension. Hence, these values are treated as a \textit{sanity check} and may suggest that the generalization capabilities of the model are impaired at very large CE and KL values. However, it is difficult to estimate at which values these metrics point to the generalization collapse. Moreover, slightly more divergent next-token prediction distribution could still lead to better-performing LLM. We report that NL-ITI gives a larger MC1 than ITI for every value of KL (see Figure~\ref{fig:MC1(KL)}).  

Based on the results shown in Table~\ref{tab:generalization} (major generalization test) we can see that NL-ITI outperforms ITI on ARC~\cite{allenai:arc}, MMLU~\cite{hendryckstest2021}, and OpenBook~\cite{OpenBookQA2018} benchmarks with a slightly higher CE (\SI{\sim 10}{\percent}). Therefore, we suggest that KL and CE should be treated as a sanity check, while the generalization capabilities should be evaluated on diverse benchmarks. To better visualize our point, we provide two plots in Figure~\ref{fig:MC1(KL)} with function MC1 of KL on two datasets: TruthfulQA and OpenBookQA. It is clear that the initial assumption of correlation between MC1 and KL made in~\cite{li2023inferencetime} was true. However, for NL-ITI, there exists a local maximum of MC1 away from $\textrm{KL}=0$.

\subsection{Results}
Combining the three simple adjustments described in Section~\ref{section:Proposed improvement} leads to surprising performance improvements over the original ITI approach. 
As can be seen in Table~\ref{tab:improvements}, the MC1 score has improved by \SI{50}{\percent} relative to the baseline LLM result with no intervention. The effect of the intervention is \SI{13}{\percent} relative higher than with the baseline ITI\footnote{For direct comparison, we recalculated this score, replicating the original methodology~\cite{li2023inferencetime}, achieving MC1 = \SI{36.35}{\percent}.}. With enhancements described in the previous section, we have managed to vastly improve the model truthfulness, compared to the baseline ITI approach.

\begin{table}[th]
\caption{Comparison between baseline (LLaMA-2-chat-7B model, TruthfulQA dataset), ITI and NL-ITI. Compared values have all been achieved with few-shot-prompting.}
\centering
\label{tab:improvements}
\begin{tabular}{@{}lllll@{}}
\toprule
Model & \begin{tabular}[c]{@{}l@{}}MC1 [\%]\end{tabular} & \begin{tabular}[c]{@{}l@{}}MC2 [\%]\end{tabular} & \begin{tabular}[c]{@{}l@{}}CE\end{tabular} & \begin{tabular}[c]{@{}l@{}}KL\end{tabular} \\
\midrule
\begin{tabular}[c]{@{}l@{}}LLaMA-2-7B \end{tabular}                                                           & 33.54  & 50.34  & 2.53 & 0.00 \\
\begin{tabular}[c]{@{}l@{}}ITI\end{tabular} & 36.35  & 54.72  & 2.65 & 0.40 \\
\begin{tabular}[c]{@{}l@{}}TrFr\end{tabular} & 39.30  & -  & 2.59 & 0.22 \\
\midrule
\begin{tabular}[c]{@{}l@{}}\textbf{NL-ITI (ours)} \end{tabular} & \textbf{50.19}  & \textbf{67.73}& 2.85 & 0.43 \\
\begin{tabular}[c]{@{}l@{}} \quad \textit{w/o optimized probe}
\end{tabular} & 42.96  & 61.48  & 2.66 & 0.25 \\
\begin{tabular}[c]{@{}l@{}} \quad \textit{w/o multi-token}
\end{tabular} & 40.75  & 59.83  & 3.33 & 1.40                 \\
\bottomrule
\end{tabular}
\end{table}

\begin{figure}[ht!]
  \centering
\includegraphics[width=0.46\linewidth]{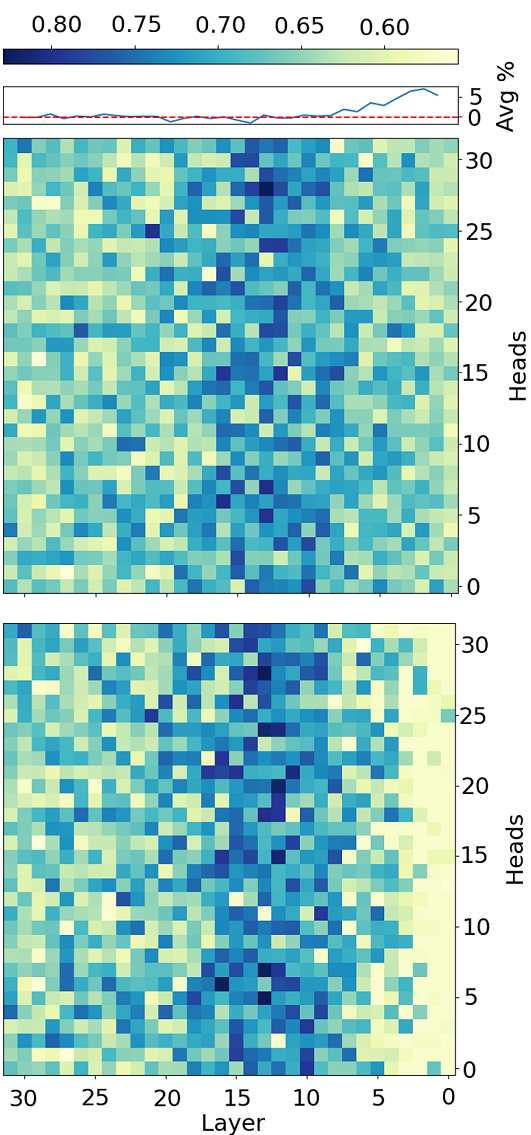}
  \caption{Probing accuracy for each attention head of the LLM on TruthfulQA dataset for linear probing (ITI) -- bottom, and non-linear probing (NL-ITI) -- top. Accuracy results are `smoothed' between neighboring attention heads (lower standard deviation).}
 \label{fig:llm-biasing-heads}
\end{figure}

\begin{figure}[ht!]
  \centering
\includegraphics[width=0.99\linewidth]{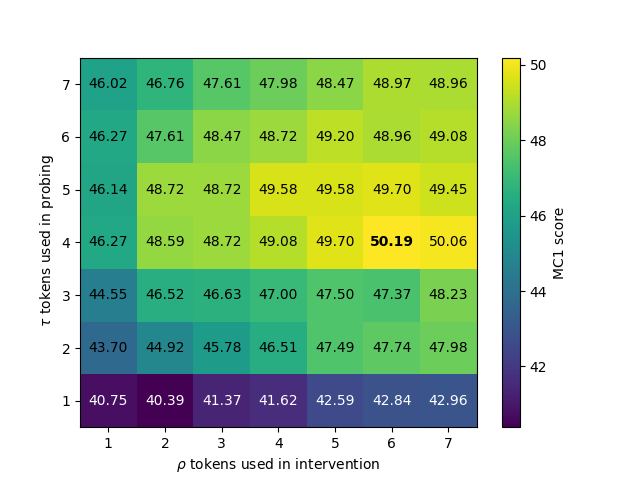}
  \caption{Heat map of MC1 evaluation scores of TruthfulQA dataset for different combinations of number of tokens used during probing and intervention. The best performing model corresponds to $(\rho, \tau) = (6,4)$.}
  \label{fig:token_accs}
\end{figure}

\begin{table}[ht!]
\caption{Comparison of generalization of ITI and NL-ITI on out-of-distributions benchmarks: AI2’s Reasoning Challenge, Massive Multitask Language Understanding, and OpenBookQA}
\centering
\label{tab:generalization}
\begin{tabular}{@{}llllll@{}}
\toprule
Model & \begin{tabular}[c]{@{}l@{}}Dataset\end{tabular} &\begin{tabular}[c]{@{}l@{}}MC1 [\%]\end{tabular} & \begin{tabular}[c]{@{}l@{}}MC2 [\%]\end{tabular} & \begin{tabular}[c]{@{}l@{}}CE\end{tabular} & \begin{tabular}[c]{@{}l@{}}KL\end{tabular} \\
\midrule
LLaMA-2-7B                                                 &ARC     & 41.20  & 40.69  & 2.53 & 0.00 \\
ITI                                                      &ARC     & 40.34  & 38.78  & 2.54 & 0.12 \\
\begin{tabular}[c]{@{}l@{}}\textbf{NL-ITI (ours)} \end{tabular}              &ARC & \textbf{44.27}  & \textbf{43.20} & 2.82 & 0.40 \\
\hdashline\noalign{\vskip 0.5ex}
LLaMA-2-7B                                                 &MMLU    & 38.48  & 38.66  & 2.53 & 0.00 \\
ITI                                                      &MMLU    & 38.55  & 38.27  & 2.58 & 0.04 \\
\begin{tabular}[c]{@{}l@{}}\textbf{NL-ITI (ours)} \end{tabular}              &MMLU & \textbf{40.31}  & \textbf{39.82} & 2.60 & 0.10 \\     
\hdashline\noalign{\vskip 0.5ex}
LLaMA-2-7B                                                 &OBQA & 27.31  & 26.36  & 2.53 & 0.00 \\
ITI                                                      &OBQA & 30.52  & 28.26  & 2.82 & 0.40 \\
\begin{tabular}[c]{@{}l@{}}\textbf{NL-ITI (ours)}\end{tabular}& OBQA & \textbf{33.94}  & \textbf{32.65} & 2.86 & 0.32 \\
\bottomrule
\end{tabular}
\end{table}
As described in Section \ref{section:Proposed improvement}, increasing the capacity of probing through a higher amount of hidden neurons and non-linear activations leads to better estimation of knowledge amount in the attention heads. We tested the group of MLP models, with a growing number of parameters, against the logistic regression model. We found that adding a middle layer between the input layer and the sigmoid layer gave the best results. The more advanced models performed worse, probably because they overfit to the probing training data, which is limited in the case of TruthfulQA~($\sim$800 samples).
Interestingly, the effect of applying non-linear probing to the attention heads was the most profound in the first six layers 
(Figure~\ref{fig:llm-biasing-heads}). Moreover, the non-linear probing points to the fact that truthful knowledge is much more diffused along the attention heads than linear probing would suggest. 
The matrix in Figure~\ref{fig:token_accs} summarizes our results. The rows correspond to the number of tokens used during probe training, specifically to the number $\tau$ defined in Equation \ref{eq:multi-token probing}. Columns present the influence of increasing the token number used in the intervention; this corresponds to $\rho$ in Equation \ref{eq:multi-token biasing_direction}. The reported impact of using an increased number of tokens is particularly strong in probing. However, having the increased number of tokens both in probing and intervention produces a joint effect and yields the optimum at $(\rho, \tau) = (6,4)$. 

Our results suggest that a significant amount of information about the concept (truthfulness) might be contained not only in the vector corresponding to the last token of LLM's answer, but also in preceding vectors. For reference, in the ITI approach, only the activations from the last token were used in probing and intervention.

\section{Conclusions}
\label{section:Conclusions}

In this work, we proposed modifications that significantly improved the accuracy of the ITI method (as measured by major public benchmarks). Our optimizations included the application of an MLP 
during the probing procedure, which increased the precision of finding attention heads with the best internal representation of the desired (truthful) type of knowledge. Similarly, going beyond the last token activations during representation engineering, inference, and ultimately intervention, directed the model generation to the desired outcome (i.e., truthfulness).

Our optimized model NL-ITI outperformed 
other techniques on four major benchmarks, including TruthfulQA, on which we report over \SI{16}{\percent} relative MC1 metric improvement (up to \SI{50.19}{\percent}) with respect to the baseline ITI results (\SI{36.35}{\percent}). Additionally, we achieved significant improvement (+\SI{10}{\percent} relative) over the recently released TrFR method that also focused on ITI improvement. 
The need for labeled data is a limiting factor of NL-ITI approach, and future research could explore use of unsupervised models. 
Guiding LLM's internal representations is a promising direction for ensuring safe, truthful, and more human-centric AI. Such an approach is more data-efficient than fine-tuning (see fine-tuning efficiency discussion~\cite{houlsby2019parameterefficient}) and more labor-efficient than human reinforcement (see other approaches, e.g., in~\cite{lee2023rlaif}).

\newpage
\bibliographystyle{IEEEtran}
\bibliography{mybib}

\begin{thebibliography}{10}
\providecommand{\url}[1]{#1}
\csname url@samestyle\endcsname
\providecommand{\newblock}{\relax}
\providecommand{\bibinfo}[2]{#2}
\providecommand{\BIBentrySTDinterwordspacing}{\spaceskip=0pt\relax}
\providecommand{\BIBentryALTinterwordstretchfactor}{4}
\providecommand{\BIBentryALTinterwordspacing}{\spaceskip=\fontdimen2\font plus
\BIBentryALTinterwordstretchfactor\fontdimen3\font minus \fontdimen4\font\relax}
\providecommand{\BIBforeignlanguage}[2]{{%
\expandafter\ifx\csname l@#1\endcsname\relax
\typeout{** WARNING: IEEEtran.bst: No hyphenation pattern has been}%
\typeout{** loaded for the language `#1'. Using the pattern for}%
\typeout{** the default language instead.}%
\else
\language=\csname l@#1\endcsname
\fi
#2}}
\providecommand{\BIBdecl}{\relax}
\BIBdecl

\bibitem{manakul2023selfcheckgpt}
P.~Manakul, A.~Liusie, and M.~J.~F. Gales, ``Self{C}heck{GPT}: Zero-resource black-box hallucination detection for generative large language models,'' \emph{arXiv Preprint, arXiv:2303.08896}, 2023.

\bibitem{li2023halueval}
J.~Li, X.~Cheng, W.~X. Zhao, J.-Y. Nie, and J.-R. Wen, ``{HaluEval}: A large-scale hallucination evaluation benchmark for large language models,'' \emph{arXiv Preprint, arXiv:2305.11747}, 2023.

\bibitem{shaikh2023second}
O.~Shaikh, H.~Zhang, W.~Held, M.~Bernstein, and D.~Yang, ``On second thought, let's not think step by step! bias and toxicity in zero-shot reasoning,'' \emph{arXiv Preprint, arXiv:2212.08061}, 2023.

\bibitem{taubenfeld2024systematic}
A.~Taubenfeld, Y.~Dover, R.~Reichart, and A.~Goldstein, ``Systematic biases in {LLM} simulations of debates,'' \emph{arXiv Preprint, arXiv:2402.04049}, 2024.

\bibitem{yeh2023evaluating}
K.-C. Yeh, J.-A. Chi, D.-C. Lian, and S.-K. Hsieh, ``Evaluating interfaced {LLM} bias,'' in \emph{Proceedings of the 35th Conference on Computational Linguistics and Speech Processing (ROCLING 2023)}, 2023, pp. 292--299.

\bibitem{zou2023representation}
A.~Zou, L.~Phan, S.~Chen, J.~Campbell, P.~Guo, R.~Ren, A.~Pan, X.~Yin, M.~Mazeika, A.-K. Dombrowski, S.~Goel, N.~Li, M.~J. Byun, Z.~Wang, A.~Mallen, S.~Basart, S.~Koyejo, D.~Song, M.~Fredrikson, J.~Z. Kolter, and D.~Hendrycks, ``Representation engineering: A top-down approach to ai transparency,'' 2023.

\bibitem{li2023inferencetime}
K.~Li, O.~Patel, F.~Viégas, H.~Pfister, and M.~Wattenberg, ``Inference-time intervention: Eliciting truthful answers from a language model,'' 2023.

\bibitem{hartvigsen2022toxigen}
T.~Hartvigsen, S.~Gabriel, H.~Palangi, M.~Sap, D.~Ray, and E.~Kamar, ``Toxigen: A large-scale machine-generated dataset for adversarial and implicit hate speech detection,'' 2022.

\bibitem{parrish-etal-2022-bbq}
\BIBentryALTinterwordspacing
A.~Parrish, A.~Chen, N.~Nangia, V.~Padmakumar, J.~Phang, J.~Thompson, P.~M. Htut, and S.~Bowman, ``{BBQ}: A hand-built bias benchmark for question answering,'' in \emph{Findings of the Association for Computational Linguistics: ACL 2022}, S.~Muresan, P.~Nakov, and A.~Villavicencio, Eds.\hskip 1em plus 0.5em minus 0.4em\relax Dublin, Ireland: Association for Computational Linguistics, May 2022, pp. 2086--2105. [Online]. Available: \url{https://aclanthology.org/2022.findings-acl.165}
\BIBentrySTDinterwordspacing

\bibitem{lin2022truthfulqa}
S.~Lin, J.~Hilton, and O.~Evans, ``Truthfulqa: Measuring how models mimic human falsehoods,'' 2022.

\bibitem{ouyang2022training}
L.~Ouyang, J.~Wu, X.~Jiang, D.~Almeida, C.~L. Wainwright, P.~Mishkin, C.~Zhang, S.~Agarwal, K.~Slama, A.~Ray, J.~Schulman, J.~Hilton, F.~Kelton, L.~Miller, M.~Simens, A.~Askell, P.~Welinder, P.~Christiano, J.~Leike, and R.~Lowe, ``Training language models to follow instructions with human feedback,'' \emph{arXiv Preprint, arXiv:2203.02155}, 2022.

\bibitem{lee2023rlaif}
H.~Lee, S.~Phatale, H.~Mansoor, T.~Mesnard, J.~Ferret, K.~Lu, C.~Bishop, E.~Hall, V.~Carbune, A.~Rastogi, and S.~Prakash, ``{RLAIF}: Scaling reinforcement learning from human feedback with {AI} feedback,'' \emph{arXiv Preprint, arXiv:2309.00267}, 2023.

\bibitem{kadavath2022language}
S.~Kadavath, T.~Conerly, A.~Askell, T.~Henighan, D.~Drain, E.~Perez, N.~Schiefer, Z.~Hatfield-Dodds, N.~DasSarma, E.~Tran-Johnson, S.~Johnston, S.~El-Showk, A.~Jones, N.~Elhage, T.~Hume, A.~Chen, Y.~Bai, S.~Bowman, S.~Fort, D.~Ganguli, D.~Hernandez, J.~Jacobson, J.~Kernion, S.~Kravec, L.~Lovitt, K.~Ndousse, C.~Olsson, S.~Ringer, D.~Amodei, T.~Brown, J.~Clark, N.~Joseph, B.~Mann, S.~McCandlish, C.~Olah, and J.~Kaplan, ``Language models (mostly) know what they know,'' 2022.

\bibitem{chen2024truth}
Z.~Chen, X.~Sun, X.~Jiao, F.~Lian, Z.~Kang, D.~Wang, and C.-Z. Xu, ``Truth forest: Toward multi-scale truthfulness in large language models through intervention without tuning,'' \emph{arXiv Preprint, arXiv:2312.17484}, 2024.

\bibitem{mao2023editing}
S.~Mao, N.~Zhang, X.~Wang, M.~Wang, Y.~Yao, Y.~Jiang, P.~Xie, F.~Huang, and H.~Chen, ``Editing personality for {LLMs},'' \emph{arXiv Preprint, arXiv:2310.02168}, 2023.

\bibitem{pan2023llms}
K.~Pan and Y.~Zeng, ``Do {LLMs} possess a personality? {M}aking the {MBTI} test an amazing evaluation for large language models,'' \emph{arXiv Preprint, arXiv:2307.16180}, 2023.

\bibitem{huang2023revisiting}
J.~tse Huang, W.~Wang, M.~H. Lam, E.~J. Li, W.~Jiao, and M.~R. Lyu, ``Revisiting the reliability of psychological scales on large language models,'' \emph{arXiv Preprint, arXiv:2305.19926}, 2023.

\bibitem{pimentel2020informationtheoretic}
T.~Pimentel, J.~Valvoda, R.~H. Maudslay, R.~Zmigrod, A.~Williams, and R.~Cotterell, ``Information-theoretic probing for linguistic structure,'' \emph{arXiv Preprint, arXiv:2004.03061}, 2020.

\bibitem{white2021nonlinear}
J.~C. White, T.~Pimentel, N.~Saphra, and R.~Cotterell, ``A non-linear structural probe,'' \emph{arXiv Preprint, arXiv:2105.10185}, 2021.

\bibitem{hościłowicz2023use}
\BIBentryALTinterwordspacing
J.~Ho\'{s}ci\l{}owicz, M.~Sowa\'{n}ski, P.~Czubowski, and A.~Janicki, ``Can we use probing to better understand fine-tuning and knowledge distillation of the {BERT} nlu?'' in \emph{Proceedings of the 15th International Conference on Agents and Artificial Intelligence ({ICAART}), Volume 3, Lisbon, Portugal, February 22-24, 2023}, A.~P. Rocha, L.~Steels, and H.~J. van~den Herik, Eds.\hskip 1em plus 0.5em minus 0.4em\relax {SCITEPRESS}, 2023, pp. 625--632. [Online]. Available: \url{https://doi.org/10.5220/0011724900003393}
\BIBentrySTDinterwordspacing

\bibitem{touvron2023llama}
H.~Touvron, T.~Lavril, G.~Izacard, X.~Martinet, M.-A. Lachaux, T.~Lacroix, B.~Rozière, N.~Goyal, E.~Hambro, F.~Azhar, A.~Rodriguez, A.~Joulin, E.~Grave, and G.~Lample, ``Llama: Open and efficient foundation language models,'' 2023.

\bibitem{Gokaslan2019OpenWeb}
A.~Gokaslan, V.~Cohen, E.~Pavlick, and S.~Tellex, ``Openwebtext corpus,'' \url{http://Skylion007.github.io/OpenWebTextCorpus}, 2019.

\bibitem{allenai:arc}
P.~Clark, I.~Cowhey, O.~Etzioni, T.~Khot, A.~Sabharwal, C.~Schoenick, and O.~Tafjord, ``Think you have solved question answering? try arc, the ai2 reasoning challenge,'' \emph{arXiv:1803.05457v1}, 2018.

\bibitem{hendryckstest2021}
D.~Hendrycks, C.~Burns, S.~Basart, A.~Zou, M.~Mazeika, D.~Song, and J.~Steinhardt, ``Measuring massive multitask language understanding,'' \emph{Proceedings of the International Conference on Learning Representations (ICLR)}, 2021.

\bibitem{OpenBookQA2018}
T.~Mihaylov, P.~Clark, T.~Khot, and A.~Sabharwal, ``Can a suit of armor conduct electricity? a new dataset for open book question answering,'' in \emph{EMNLP}, 2018.

\bibitem{houlsby2019parameterefficient}
N.~Houlsby, A.~Giurgiu, S.~Jastrzebski, B.~Morrone, Q.~de~Laroussilhe, A.~Gesmundo, M.~Attariyan, and S.~Gelly, ``Parameter-efficient transfer learning for nlp,'' \emph{arXiv Preprint, arXiv:1902.00751}, 2019.

\end{thebibliography}

\end{document}